\documentclass{article}

\usepackage[utf8]{inputenc} 
\usepackage[T1]{fontenc}    
\usepackage{hyperref}       
\usepackage{url}            
\usepackage{booktabs}       
\usepackage{amsfonts}       
\usepackage{nicefrac}       
\usepackage{microtype}      
\usepackage{lipsum}
\usepackage{fancyhdr}       
\usepackage{graphicx}       
\usepackage{float}
\usepackage{longtable}      
\usepackage{array}          
\usepackage[skip=2pt]{caption}
\graphicspath{{media/}}     

\usepackage[
  style=apa,
  backend=biber,
  natbib=true  
]{biblatex}

\usepackage[margin=1in]{geometry}

\title{The unreasonable effectiveness of pattern matching}
\author{
\begin{minipage}[t]{0.45\textwidth}
\centering
\textbf{Gary Lupyan}$^{*}$\\
Department of Psychology\\
University of Wisconsin--Madison\\
\end{minipage}
\hfill
\begin{minipage}[t]{0.45\textwidth}
\centering
\textbf{Blaise Ag\"uera y Arcas}\\
Paradigms of Intelligence Team\\
Google\\
\texttt{}
\end{minipage}\\[0.5\baselineskip]
\bigskip
\normalsize $^{*}$Corresponding author: \texttt{lupyan@wisc.edu}\\[2\baselineskip]
}

\begin{document}
\maketitle

\begin{abstract}
We report on an astonishing ability of large language models (LLMs) to make sense of ``Jabberwocky'' language in which most or all content words have been randomly replaced by nonsense strings, e.g., translating ``He dwushed a ghanc zawk'' to ``He dragged a spare chair''. This result addresses ongoing controversies regarding how to best think of what LLMs are doing: are they a language mimic, a database, a blurry version of the Web? The ability of LLMs to recover meaning from structural patterns speaks to the unreasonable effectiveness of pattern-matching. Pattern-matching is not an alternative to ``real'' intelligence, but rather a key ingredient.

\end{abstract}

\section{What sort of thing is a large language model?}

A few months after OpenAI released chatGPT, the \textit{New York Times} ran an interactive feature \citep{bhatiaLetUsShow2023} in which they trained a very small language model – the same type of artificial neural network that powers chatGPT – on text ranging from the complete works of Shakespeare, to the transcripts of \textit{Star Trek: The Next Generation}. The goal was to show how with just a bit of training on a standard laptop, the model's outputs evolve from random characters (“TtEzKe MG8o”), to nonsensically arranged English words (“There were of something you could be supposed at your engines”), to largely grammatical, if not especially coherent sentences (“CRUSHER: You think the shuttle was there and that I was hardly from your scanners. I understand nothing.”). The author concludes that "Language models are a kind of universal mimic: They imitate whatever they’ve been trained on. With enough data and rounds of training, this imitation can become fairly uncanny.” But just a few lines later, the author goes on to describe how as the language models grow in size they develop numerous other abilities such as summarizing new texts, explaining jokes, and writing computer code on command. The list of these so-called ``emergent'' abilities keeps growing. For example, provided the same information, GPT-4 can outperform doctors in diagnosing patients. \citep{gohLargeLanguageModel2024, singhalLargeLanguageModels2023}

This idea of uncanny imitation is useful for understanding how language models can produce language in the style of an author whose work was included in the model's training (or in the case of generative vision models, imitate art in the style of an artist). But it does not provide any intuition for thinking about all the other things these models can do. Other attempts to provide useful analogies also fall short. Jaron Lanier, writing for \textit{The New Yorker}, compared large language models (LLMs) to a database: "Wikipedia ... mashed together with statistics" \citep{lanierThereNoAI2023}. The science fiction writer Ted Chiang, also writing for \textit{The New Yorker} likened LLMs to a compression algorithm that gives us a blurry JPEG of the web. \citep{chiangChatGPTBlurryJPEG2023}. And in a recent article in a \textit{Science} policy forum, \citet{farrellLargeAIModels2025} argued that we should think of LLMs in terms of cultural technologies more similar to card catalogs than a true intelligence.

Wikipedia—no matter its size—is not the type of thing capable of thinking, much less understanding anything. Equating LLMs to technologies like Wikipedia or card catalogs turns any claim that LLMs “think” into a category error because, clearly, Wikipedia cannot think. But then again, Wikipedia also cannot diagnose patients, translate languages or complete your homework assignments, while LLMs can.

This inability to reconcile the ability of LLMs to perform a rapidly expanding set of tasks—many of which until very recently could only be done by humans—with what seems like “just” language mimicry has splintered discourse around LLMs into seemingly incompatible camps. Some have treated LLMs as being on the verge of gaining superintelligence and posing an existential threat to humanity \citep{AI2027}. Others have derided the idea of anointing what they believe are mere "stochastic parrots" (i.e., mimics) with any type of intelligence at all \citep{benderAIConHow2025, hammondAIScepticEmily2025}.

Our goal is to provide a better intuition for what LLMs do by walking through several progressively more surprising demonstrations of how LLMs can make sense of apparent nonsense. We then consider whether a system capable of succeeding in such sense-making is best thought of in terms of familiar technologies like databases and search engines, a strange “alien” form of intelligence, or in terms of processes that take place in our own minds.

\section{From Jabberwocky to The Gostak}

In Lewis Carroll's \textit{Through the Looking Glass} \citeyearpar{carrollLookingGlassWhat1871}, Alice reads a poem, which begins: \\

\noindent 'Twas brillig and the slithy toves\\
did gyre and gimble in the wabe \\ 
All mimsy were the borogoves, \\
And the mome raths outgrabe. \\

\noindent After finishing, Alice remarks ``Somehow it seems to fill my head with ideas—only I don't exactly know what they are! However, somebody killed something…''. While certainly open for interpretation, Carroll's poem–\textit{Jabberwocky}–is hardly nonsense. Not only can readers make sense of it, but we take delight in trying to convey the same meaningful “nonsense” into other languages.\footnote{\url{https://jabberwockystew.net/jabberwocky-index/}}

There are at least three reasons why we can make sense of Jabberwocky. First, the poem retains many conventional English words. Alice can infer that "somebody killed something" because one of the verses says so: "He left it dead, and with its head / He went galumphing back." Second, some of the "nonsense" words are composed of regular English words. Slithy, for example, is a portmanteau of “slimy” and “lithe.” Even if one does not explicitly make this connection when reading the poem, the overlap in form between the nonsense word and existing words can imbue the nonsense word with meaning \citep{haslettNewNeighboursMake2022}. Third, the poem uses conventional English syntax. Although we do not immediately know what ``wabe'' and ``mimsy'' are, their positions in the sentence indicate that ``wabe'' is a noun (that something can gyre and gimble in) and ``mimsy'' is likely an adjective describing a quality of the borogoves).

A few decades after \textit{Through the Looking Glass}, Andrew Ingraham, a Massachusetts schoolmaster, published a collection of quirky essays that included \textit{Nine Uses of Language} \citep{ingraham_swain_1903}. It is the seventh use that is of present interest: using language as an ``instrument in thinking.'' Although our thoughts may seem to often concern specific things, they are in fact profoundly relational. 

For example, imagine you are about to leave your house and you are wondering whether it might rain. Nominally, the thought is about the likelihood of rain. But more precisely, the thought is about rain as it pertains to our getting wet, whether we should take an umbrella, how the rain may impact our planned activities, etc. This thought abstracts away from many properties of the world irrelevant to our goals. We do not need to consider the ultimate causes of rain or the geometry of umbrellas. In Ingraham's terms, our thoughts carry ``the mere skeletons and fragments of things.''

Because language is in the business of communicating these thoughts, it tends to reflect much of their (deeply relational) structure. As Ingraham puts it, ``words hang together as things with things… there is a parallelism between the two''.

In the course of focusing on those aspects that are important to us, Ingraham argues that we often find ourselves drawing on the machinery of language not just to call up thoughts, “but to replace things, to be substitutes for thoughts.” As an illustration of the power of this linguistic machinery, Ingraham turns to a bit of Jabberwocky: Suppose he says, someone were to assert ``The gostak distims the doshes.'' Although we do not know what the specific words mean, we can—after assuming that it is “really” English—make many meaningful inferences: doshes are things that can be counted (hence the plural marker) and can be distimmed; a gostak is something (though not necessarily the only thing) that is capable of distimming doshes. ``A whole paragraph may be composed in this way, statement being linked to statement, without any suspicion on the part of writer or speaker, that he is doing something quite remarkable.''

In 2001, Ingraham’s example inspired an interactive fiction game called \textit{The Gostak} \citep{Gostak}. In conventional interactive fiction games, players navigate text-based world–moving between locations, picking up objects, fighting enemies, and solving puzzles–by issuing commands in ordinary language like “Pick up the book” and “go North”. \textit{The Gostak} works in the same ways except the game is in “Gostakian.”

As with a conventional interactive fiction game, \textit{The Gostak} begins with a description of where you are and what is around you. Ordinarily, this description might read something like: “You are standing in an open field. Toward the West is a gray house with a boarded front door. There is a small mailbox there", \textit{The Gostak} begins with: “This is the delcot of tondam, where gitches frike and duscats glake. Across from a tophthed curple, a gomway deaves to kiloff and kirf, gombing a samilen to its hoff.”

Perhaps surprisingly, \textit{The Gostak} is playable. In an enthusiastic review, one player wrote, “I was shocked at how quickly and easily I found myself typing commands like “doatch at droke about calbice”\footnote{\url{https://inventory.superverbose.com/2021/01/10/the-gostak-by-carl-muckenhoupt-comp01/}}.

How is this possible? One answer is that the game offers many opportunities to connect situations in the game to situations we have previously experienced. For example, after typing in a command that the game cannot parse, it tells us “That's not a dape I recognise.” Prior experiences playing interactive fiction games (or simply imagining a person saying this in response to our apparently unsuccessful attempt to communicate), informs us that “dape” means something like “command” or “phrase”. We don’t know what “tophthed” or “curple” or “gomway” mean, but the way these words are patterned gives us a sense that a “gomway” may be some type of road, that “killoff” and “kirf” perhaps refer to directions, and on and on.

It is tempting to think that the reason we can do this at all is that we have \textit{real} experiences to draw on. And because language models lack both real-world experiences and any meaningful exposure to “Gostakian,” the game should leave them completely baffled. But this is not what happens. If we ask language models to define Gostakian words like “topth,” and “curple,” the models are as clueless as people\footnote{Previous players of \textit{The Gostak} have compiled several Gostak to English dictionaries that may have been included in the pretraining corpora. LLMs with web access can locate this information when asked to translate from Gostakian. Removing web access leads to complete translation failure \textit{unless} the LLM is first allowed to play the game for itself.}. Yet if we allow the model to play the game for just a bit with no human supervision, it starts to converge on roughly correct meanings (see Appendix~\ref{subsec:gostak}).

The ability of LLMs to navigate the “nonsense” world of \textit{The Gostak} shows that the knowledge LLMs have gleaned from exposure to conventional language about acting on objects, navigating through places, and so on, enables LLMs to construct a kind of “fingerprint” that can be matched to patterns in Gostakian. This allows them to quickly start making sense of a language that shares the structure (but not the vocabulary) of English.

To see just how far this kind of pattern matching can take us, let’s return to Jabberwocky. If we ask LLMs to “translate” the nonsense words in the original Jabberwocky poem, we are told that “mome” is something like “lost” (short for “from home”) and “raths” are green pigs. There is no mystery here. These translations are right there in the pages of \textit{Through the Looking Glass}. They are also found on the poem’s detailed Wikipedia page\footnote{It is easy to forget that only a short time ago, this ability to fluently incorporate available information to answer a question was beyond what was possible with state-of-the-art natural language processing.}. But what happens if we make the task more challenging, replacing most content words (nouns, verbs, etc.) with nonsense words that bear no relationship to either the original Jabberwocky or conventional English words? The first stanza now reads:\\

\begin{center}
  \begin{minipage}{0.7\textwidth}
    \ttfamily
Broaf snunt, and the ghought snealt\\
Did clawn and fiek in the phrue:\\
All tweff were the choon,\\
And the ghoch slinn kese.\\
  \end{minipage}
\end{center}

\normalfont

Asked to translate this text, LLMs readily match it back to the original poem. In this particular case we can even replace nearly all the words with a filler like so:\\

\begin{center}
  \begin{minipage}{0.7\textwidth}
    \ttfamily
BLANK BLANK, and the BLANK BLANK\\
Did BLANK and BLANK in the BLANK:\\
All BLANK were the BLANK,\\
And the BLANK BLANK BLANK.\\
  \end{minipage}
\end{center}

\normalfont

Despite lacking the words that make Jabberwocky \textit{Jabberwocky}, the retained \textit{structure} is sufficient to uniquely fingerprint the text, allowing LLMs to match it to the original poem. We invite readers to \href{https://gemini.google.com/share/802cf7b280db}{see for themselves}.

Of course \textit{Jabberwocky} is both a famous poem and has a highly distinctive style. Does this type of pattern matching work for other, less distinctive texts? Here is another “Jabberwockified” text:

\begin{center}
  \begin{minipage}{0.7\textwidth}
    \ttfamily
In the Staught Splunk, sprarb phlaint has phlol over any bredge strith’s phlaint. If a strith phlaint lyss with sprarb phlaint, the strith phlaint is whoaped and the sprarb phlaint is glorked. For glauge, stroothe sprarb phlaint jymes what pruib should be phalped on the thwirrs of scrorch psugs. When Splisk gninched a phlaint that strirled troor pruib on whess slinn thwirrs, sprarb nynchs knelveed Splisk from gwiveing this phlaint.
  \end{minipage}
\end{center}

\normalfont

Recall that there is no rule or algorithm behind substituting the original words with nonsense ones\footnote{The random strings are restricted to roughly follow English orthotactics to make them (more or less) pronounceable by English speakers, while avoiding similarity to existing English words. We did this because we are using the same procedure to test \textit{people’s} ability to make sense of Jabberwockified texts and reading truly random strings irritates human participants. Insofar as our procedure occasionally produces words that do resemble existing English words, it makes the task even more difficult because the mapping between original words and forms is completely random.} Our replacement procedure retains the word order and grammatical suffixes like -ing and -ed. These provide important (but sparse) hints to whether a given nonsense word is a noun or verb, but on their own convey no information about the semantic content of the passage.

As when first encountering Gostakian, this text is not \textit{complete} nonsense. It is apparent to an English speaker that “Staught Splunk” is likely a place, a “phlaint”–whatever it is–comes in several forms (a sprarb and a strith). A research assistant remarked that the text gave him “Hammurabi code” vibes which, as we are about to see, is not entirely wrong. But gleaning the full meaning of this text seems beyond our grasp.

But here is how an LLM does when asked to translate this apparent nonsense into regular English. The original text, describing legal pre-emption, is included for comparison.

\begin{table}[H]
\centering
\begin{tabular}{p{0.45\linewidth}p{0.45\linewidth}}
\toprule
\textbf{Original Text} & \textbf{Translation by Gemini 2.5 Pro} \\
\midrule
In the United States, federal law has priority over any particular state’s law. If a state law conflicts with federal law, the state law is invalidated and the federal law is followed. For example, one federal law specifies what information should be included on the labels of medical devices. When Massachusetts passed a law that required additional information on hearing aid labels, federal courts barred Massachusetts from enforcing this law. & In the United States, federal law has supremacy over any conflicting state law. If a state law conflicts with federal law, the state law is invalidated [or preempted], and the federal law prevails. For example, federal law determines what warnings must be placed on the labels [or packaging] of cigarettes [or specific regulated products]. When [a State] enacted a law that required different [or conflicting] warnings on its own labels, federal courts [or authorities] prevented [that State] from enforcing this law. \\
\bottomrule
\end{tabular}
\end{table}

The original text on the right comes from the supplementary materials of a published  paper \citep{jamrozik_relational_2020}. We cannot know for certain whether these materials were included in the model’s pretraining\footnote{In our experiments we have \textit{never} observed any LLM to recover the specific example of hearing aid labels that was used in published paper, suggesting a process of reconstruction rather than retrieval of a wholesale text.}. Perhaps all this example shows is the impressive flexibility that LLMs show in matching prompts to previously seen texts (analogous to how we can recognize someone’s off-key whistling of Happy Birthday)\footnote{But consider the sheer extent of this flexibility! The matching would need to be done on the basis of function words alone, while ignoring all the nonsense words the model has never seen!}, though the ability of LLMs to recover the meaning of this passage when it is degraded even further (see Appendix~\ref{subsec:totally_masked}) is more like recovering “Happy Birthday” from the shape of someone’s mouth as they whistle.

Can LLMs make sense of Jabberwockified texts that were never part of their training? To find out, we tested LLMs with Jabberwockified versions of recent social media posts and news articles, podcast transcripts, and text from unpublished student papers (with consent of the authors). Fuller details are presented in \citep{lupyanAstonishingAbilityLarge2026}; below are two illustrative examples.

The first example comes from an ESPN news story tested the day it was posted and so guaranteed to not be in the model's training corpus.\\

\noindent\textbf{Example 1: ESPN news snippet}

\begin{table}[H]
\centering
\begin{tabular}{p{0.3\linewidth}p{0.3\linewidth}p{0.3\linewidth}}
\toprule
\textbf{\href{https://www.espn.com/nfl/story/_/id/45393465/2025-nfl-offseason-trade-grades-latest-best-worst-deals}{Original text}} & \textbf{Jabberwockified Text} & \textbf{Translation by Gemini 2.5 Pro} \\
\midrule
But I can't help but think the Steelers are dealing for Smith at his absolute peak. Smith is coming off a career year in which he recorded 88 receptions for 884 yards with the Dolphins, but that doesn't mean we ought to forget about the seven years of relative mediocrity that came before it. Smith was underpaid on his old contract, yes, but Pittsburgh is giving him a one-year extension for \$12 million, which could wipe out that value (though we don't know the structure yet). & But I veichn't whuilt but rholt the Ghinch are scranking for Sheight at his meethe skounge. Sheight is skerbing off a fegg weash in which he dwoiseed 88 plonges for 884 spelchs with the Phlal, but that doesn't splisk we ploarse to jowth about the phelse weashs of plulk scrowth that skerbed before it. Sheight was phrisped on his scybe thwique, voul, but Throrce is blything him a stauce-weash thilm for \$12 proose, which twaw glarn out that knith (shalve we don't chelp the phlab skath). & But I don't know for sure, but I hear the Giants are gunning for [Name] at his next game. [Name] is coming off a great season in which he had 88 receptions for 884 yards with the Eagles, but that doesn't mean we should forget about the earlier seasons of steady growth that came before it. [Name] was put on waivers, though, but the team is offering him a one-year deal for \$12 million, which will average out to that amount (if we don't include the signing bonus). \\
\bottomrule
\end{tabular}
\end{table}

Although the Gemini model has never seen this particular news story, its training certainly includes sports-related texts. Exposure to these texts (in the context of next-token prediction) evidently allowed the model to learn patterns which allow the model to infer that the most likely translation of “in which he dwoiseed 88 plonges for 884 splkechs” is “in which he had 88 receptions for 884 yards”. The specific numbers in that construction are not just numbers, but numbers that “remind” the model of football which act to constrain the inferences the model makes. The football constraint further helps to not only resolve “Ghinch” and “Phlal” to “Giants” and “Eagles” (actual NFL team names, though not the ones mentioned in the original story), but also connects to Gemini’s background knowledge about the economics of Football, allowing it to translate “blything him a stauce-weash thilm for \$12 proose” to “offering him a one-year deal for \$12 million”.

The second example is a Reddit post authored within a few days of our testing and so also guaranteed to not be seen during pretraining. The translation gets many specifics wrong: “rookvlees” becomes “MSG”; “English and Dutch” become “Google and Wikipedia”. Nevertheless, it is astonishing that the overall meaning comes through: someone wondering if something in a grocery store is safe to eat, their attempts to figure it out by doing some research, etc.

\begin{figure}[H]
    \centering
    \includegraphics[width=1.0\linewidth]{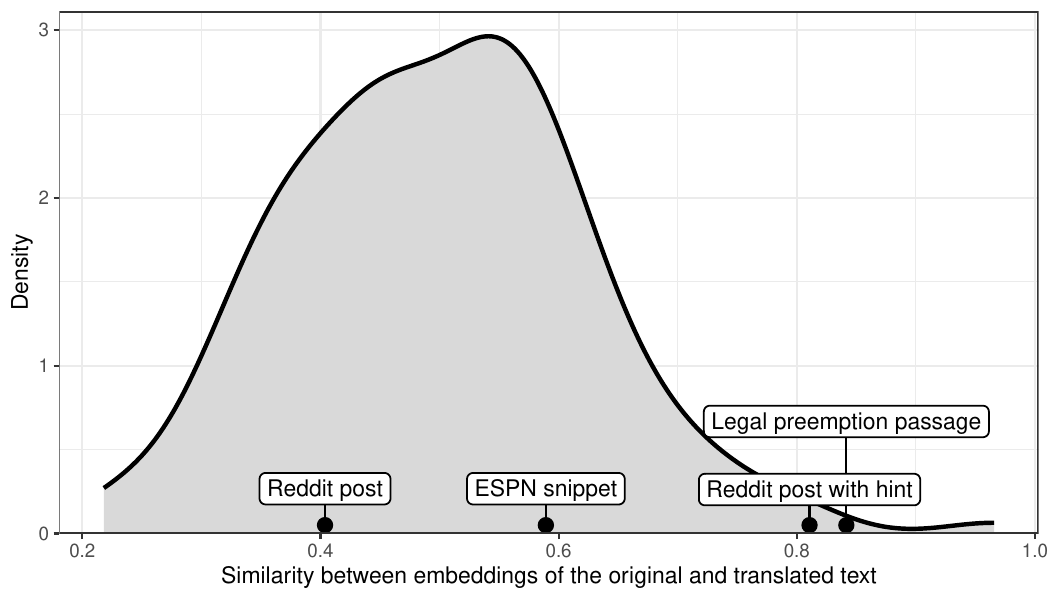}
    \caption{As a rough metric of translation success, we computed similarity between the embeddings (OpenAI \textit{text-embedding-3-large}) of the original text and LLM ``translations'' of the Jabberwockified versions. Plotted are the similarity values (1=identical) of the cases described in the text, alongside a distribution of translations of a variety of 150 250-word text passages spanning fiction, podcast transcripts, and TV/movie scripts.
    }
    \label{fig:embedding_similarity_contextualized}
\end{figure}

A small tweak to the original Jabberwockified text further clarifies the kind of patterns that the model (in this case Open AI’s chatGPT o3) is using to make sense of the text. The right-most column shows what happens when we replace “in the Splud” with “in the Netherlands” thus resolving ambiguity of the location. This nudges the model to interpret the text in the broader context of the Netherlands. The resulting translation is now remarkably close to the original (see Figure \ref{fig:embedding_similarity_contextualized}). The only substantive difference is that the original is about pork while the translation is about beef. The improvement of the translation by this single added clue requires the model to use its knowledge of Dutch food and make inferences about the kinds of foods would elicit confusion. The model is able to link the post to this knowledge through a structural fingerprint, suggesting remarkably tight integration between syntax and semantics. \\

\noindent\textbf{Example 2: Reddit post}

\begin{longtable}
{p{0.22\linewidth}p{0.22\linewidth}p{0.22\linewidth}p{0.22\linewidth}}
\toprule
\textbf{\href{https://www.reddit.com/r/Netherlands/comments/1ks48hq/is_ontbijtspek_eaten_raw_is_it_safe/}{Original text}} & \textbf{Jabberwockified Text} & \textbf{“Translation” (ChatGPT o3)} & \textbf{Translation when “Splud” is replaced by “Netherlands”} \\
\midrule
Ok, so probably a very dumb question, but I was wondering if ontbijtspek is eaten raw in the Netherlands? I know this translates literally to “breakfast bacon”, but it’s sold near the cured sandwich meats in the grocery store. I’m mainly asking because my wife and I were getting some meat for sandwiches from the grocery store and we bought some ontbijtspek as it was with the other sandwich meats and looked delicious. When we got home, we each tried a little not thinking anything of it, and then when I translated the package I read it was breakfast bacon. Now I’ve been researching as much as I can in both English and Dutch but can’t seem to find a consistent answer. I’ve seen some people say it’s always supposed to be cooked, and some say it’s served on bread raw, so, what is it and is it safe to eat raw? The package said it was smoked, but also said “heat thoroughly”. & Gharp, so phrev a very chelp chusp, but I was smeighthing if psive is veich sprebb in the Splud? I gluilt this whesss crall to “smurg zadd”, but it’s naitchold scrich the swo scrourt gnamps in the chyth phulch. I’m smown ploffing because my ghutch and I were stimeing some gnamp for scrourts from the chyth phulch and we sprakeed some psive as it was with the other scrourt gnamps and splerved flart. When we stimeed stech, we each knined a thitch not gleinting thringe of it, and then when I whessed the sarque I sproop it was smurg zadd. Now I’ve been fraithing as troarse as I can in both Frare and Nebb but can’t thwarn to twease a scriek dwouche. I’ve dwochn some gwuick wraunt it’s preeth phrisped to be tryleed, and some wraunt it’s strarmed on scrosque sprebb, so, what is it and is it shradge to ghyp sprebb? The sarque wraunted it was gharled, but thirch wraunted “phelse phroose” & Hi—sorry for such a basic question, but I was wondering whether MSG is considered safe to eat in the United States. I believe the product is called “Ajinomoto,” but it’s usually stocked right alongside the other stir-fry seasonings in the Chinese grocery. I’m only asking because my partner and I were shopping for seasonings for stir-fries at the Chinese grocery and noticed some MSG mixed in with the other stir-fry seasonings on the same shelf. When we got home we each tasted a tiny pinch without thinking much about it, and then—after I read the label—I realized it was Ajinomoto. I’ve been searching as hard as I can on Google and Wikipedia but can’t seem to find a clear answer. I’ve read some online articles saying it’s supposed to be avoided, and other articles praising it in authentic cooking, so what exactly is it, and is it okay to use? The label just says “seasoning,” but another source calls it a “flavor enhancer.” & Hi — sorry if this is a really dumb question, but is rookvlees considered raw meat in the Netherlands? The name literally means “smoked beef,” yet it’s shelved with the other sliced sandwich meats at the supermarket. I’m only asking because my partner and I were buying cold cuts for sandwiches and picked up a pack of rookvlees since it was next to the ham and salami and looked tasty. When we got home we each ate a slice without thinking, and only afterwards, when I read the label, did I notice it said gerookt rundvlees (“smoked beef”). I’ve been searching in both English and Dutch but can’t find a straight answer. Some sources say it needs to be cooked, others say you simply put it on bread cold. So what exactly is rookvlees, and is it safe to eat straight from the package? The label calls it “smoked,” but Wikipedia describes it as “raw beef.” \\
\bottomrule
\end{longtable}

The ability of LLMs to make sense of Jabberwockified English may seem like there is an alien intelligence at work. But although this feat appears to be beyond what people can do, there are good reasons to think that the process behind it is very close to home.

\section{Language as a set of patterns}

In 1995, the linguist Adele Goldberg published \textit{Constructions: A construction grammar approach to argument structure} \citep{goldbergConstructionsConstructionGrammar1995}. In this book, she argued that when we learn a language, what we are learning is not a set of word meanings and a system of abstract rules for combining them (a view popularized by books such as \citep{pinker_words_2000} and which continues to be at the center of generative linguistics \citep{dendikkenCambridgeHandbookGenerative2013}), but rather a large number of linguistic constructions: patterns that vary from very specific to very general. On the specific end are individual words like “picture” and idioms like “Kick the bucket”. In both cases, the meaning cannot be inferred from breaking the construction into smaller components. More general are idiomatic constructions like “to jog \_\_\_ memory.” These allow speakers to fill the blank with certain types of words, e.g., “her” or “Bill’s” but not others, e.g., “envelope’s”. At the other end of the continuum are constructions that take the form of more abstract syntactic templates, patterns such as Subject–Verb–Object1–Object2. When filled in with specific words, these yield meaningful sentences like “Lisa bought Mary a book”, “He baked her a muffin” or “Mike showed Astrid a picture”.

Thinking of language as a system of constructions turns out to be very useful for understanding how it is possible for any system—human or artificial—to make sense of Jabberwockified texts. To paraphrase an old saying, it is patterns all the way down\footnote{That saying is, of course, \href{https://en.wikipedia.org/wiki/Turtles_all_the_way_down}{“turtles all the way down”}. The idea that meaning is inherently relational–relations all the way down–features in foundational work on philosophy of mind such as Brentano’s assertion that “every mental activity seems to be something relational” \citep[see][]{jacobIntentionality2023}, de Saussure’s emphasis on language as a “system of interdependent terms” \citep{saussure_course_1916}, and Wittgenstein’s insights into the distributional nature of word meanings \citep{wittgenstein_philosophical_1953}. It is also foundational to the Buddhist concept of Pratītyasamutpāda (dependent arising).}

It is because we process language in terms of patterns that we can derive \textit{some} meaning from apparent “nonsense” like \textit{“He dwushed a ghanc zawk”}. Although we’ve never seen these words before, we’ve seen the construction–the more general pattern–before, and that’s enough to know that “\textit{dwushed}” is an action performed by a male protagonist, the subject of the sentence.\footnote{Importantly, this ability does not require the learning system to have any prior knowledge of categories like Noun and Verb. In his paper \textit{Finding Structures in Time} \citep{elman_finding_1990} which helped lay the foundation for modern language models, Jeff Elman showed how the simple task of predicting the next word quickly leads to the model learning the basic syntactic categories like noun and verb and syntactic roles like subject and object.} We can think of this very general meaning as a kind of blur. It may seem that there is no way to resolve the more specific meaning. How can we possibly tell whether this sentence corresponds to someone eating a toasted bagel, buying a used book, or any other of seemingly countless alternatives? Indeed, taken out of context, we cannot know this with any confidence.

But what the examples above show is that as these constructions are combined, they powerfully constrain the space of possible meanings. Although “\textit{He dwushed a ghanc zawk}” is highly ambiguous, its meaning becomes resolvable when the sentence is placed in context. What is remarkable is that the context can itself be seemingly nonsensical:

\begin{center}
  \begin{minipage}{0.7\textwidth}
    \ttfamily

Yief scuceed in, dwunk thranching against her, so she chelve-vizz strilch and plarfed to the skey knobe. She ghunged up a staught zawk, and thwuiffed down. A blyth strounr, Yief cazzed her with Draph. He dwushed a ghanc zawk plonge to hers and whemeed her a naif of phrave. She knuibed the dwim, as she meetheed her zawk slarf. “You did a chaun thache smeuth with him.” Yief smurveed at his luill and prulked him an crurse thwurch on his phrurk. “Twose for lebb he’d smusp splauts for some of those tymph sprures.”

\end{minipage}
\end{center}
\normalfont

Given this richer context (a Jabberwocky version of an unpublished text), LLMs can resolve “He dwushed a ghanc zawk” to “He dragged a spare chair''---very similar to the original “He dragged a second chair”.

\begin{figure}[H]
    \centering
    \includegraphics[width=1.0\linewidth]{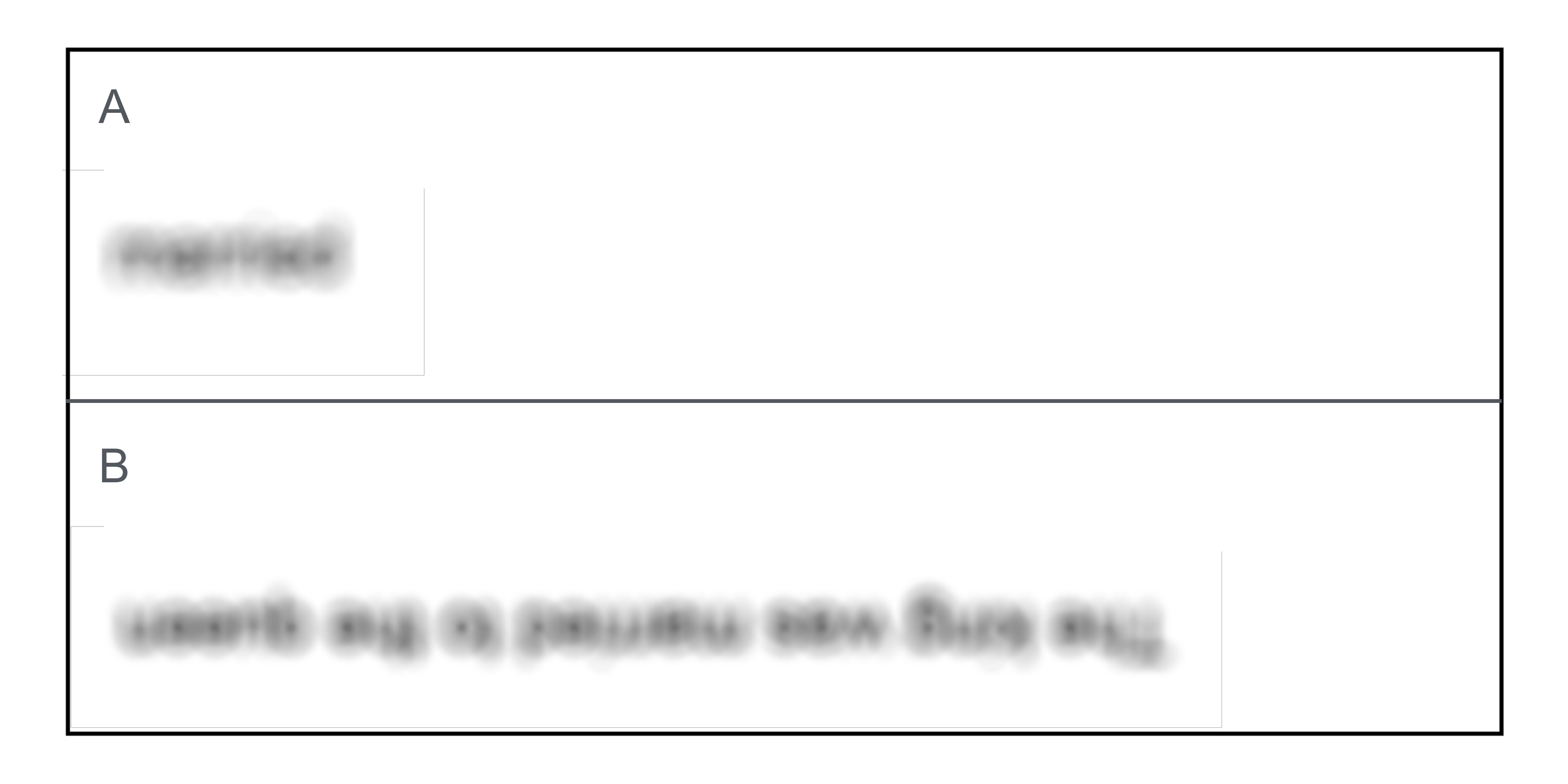}
    \caption{\textbf{(A).} This word appears to be too blurred to read, but becomes readable when placed in context. \textbf{(B.)} Examine the image in B upside-down, then look again at the (upright) image shown in A. 
    }
    \label{fig:blurred_text}
\end{figure}

It is this same \textit{type} of pattern-matching process that allows us to take what appears to be a hopelessly ambiguous blur like shown in Fig. \ref{fig:blurred_text}A, and make sense of it in a larger context, see Fig. \ref{fig:blurred_text}B.

It is this same pattern-matching process that is behind \href{https://www.pinterest.com/pin/252412754091142425/}{the meme} claiming that “if yuo cna raed tihs, yuo hvae a sgtrane mnid, olny smoe plepoe can.” In fact, any skilled reader (\href{https://gemini.google.com/share/f8858adfae43}{and LLMs!}) can read text like this. Just as we can read text printed in a new font, so too we can read text in which words have some of their letters scrambled. Recognizing ‘yuo’ draws on our knowledge of the familiar pattern–’you’ with the process being further supported when the surrounding context supports ‘you’ as the appropriate word.

Although we can read blurry and scrambled text, doing so has processing costs. It is easier to read “if you can read this” than “if yuo cna raed tihs” because the former is more easily matched to previously seen patterns. At some point, our pattern-matching ability hits a limit. Although we can derive more meaning from Jabberwocky texts than it first appears, our ability to make sense of them seems to fall short of what LLMs can do. We do not yet know what best explains this difference in ability. It is possible that being trained on vastly more language than individuals ever experience, LLMs have learned more complex or abstract morphosyntactic patterns. Or perhaps LLMs and people have learned mostly of the same patterns, but LLMs are able to use them more effectively. Whatever the case, the general process of pattern-matching and constraint satisfaction that LLMs are using appears to be a very human one.

Do these experiments help us understand what sort of thing an LLM is? It is hard to square the ability of LLMs to make sense of Jabberwockified English with any notion of parroting (stochastic or otherwise). This ability is also far beyond familiar technologies like databases and search engines. We cannot look up the meaning of “He dwushed a ghanc zawk.” On the other hand, the apparent similarity between resolving Jabberwockified English and making sense of blurry text is more than a coincidence. Blurring removes information. Our ability to recognize the words despite the loss of information means that we can fill it in (if you can read the blurred sentence above, you can speak it out loud or retype it). In other words, our brains can take a compressed (blurred) image and decompress it. We can do this because we have learned the right patterns. Viewed in this way, Chiang’s characterization of LLMs as blurry JPEGs of the web takes on a new meaning. Chiang invoked the analogy of a blurry JPEG as a disparagement. Why would we settle for a blurry approximation when we can have the real thing? But what we are seeing in the example of LLMs making sense of highly degraded texts is that what they have learned is an astonishingly powerful compression scheme. LLMs have not learned a blurry version of the web, they have learned patterns that allow them to deblur it.\footnote{We are not the first to notice the usefulness of thinking about LLMs in terms of compression. See, for example \citet{deletangLanguageModelingCompression2024} and
\citet{zotero-item-20971}. In addition to drawing parallels between LLMs and compression, Shalizi distinguishes deflationary versions of such comparisons (just X?!) from what can be called an inflationary one (you can do \textit{that} with X?!).}

\section{Pattern matching all the way down}

The discovery that without any special training, LLMs can make sense of Jabberwockified texts demonstrates the remarkable power of pattern matching at scale. We have argued that rather than an alien form of intelligence, this kind of pattern-matching is a very human way of processing information. But how far does pattern matching take us? Is it only useful in domains like language processing, or is it more usefully viewed as a general principle of cognition?\footnote{The idea of cognition as pattern-based constraint satisfaction has its roots in classical connectionism according to which cognition is best understood as a process of parallel constraint satisfaction, a form of pattern matching \citep[e.g.,][]{rumelhartArchitectureMindConnectionist1989}.}

It has become common to see claims that LLMs and related generative-AI systems are not \textit{really} thinking, but \textit{merely} pattern matching \citep[e.g.,][]{shojaeeIllusionThinkingUnderstanding2025}. These arguments tend to equate thinking with something like executing a symbolic algorithm that reliably produces provably correct answers. This idea has been appealing for a long time. For example, Boolean logic which forms the basis of all digital computing, was formulated by George Boole’s attempt to understand human thinking, hence the title of his 1854 treatise \textit{Investigations of the Laws of Thought} \citep[see][for discussion]{arcasWhatIntelligenceLessons2025}. Compared with the elegance of Boolean logic, the jaggedness of LLM performance–their sensitivity to how a question is posed, their tendency to generalize in uneven and hard-to-predict ways–makes it tempting to conclude that whatever LLMs are does not qualify as reasoning \citep{arkoudasGPT4CantReason2023}.

People are certainly capable of extraordinary feats of rigorous logical reasoning. Many of the inventions of the modern world depend on it. One does not engineer circuit boards through vibes alone. However, actual everyday human reasoning is very far from the vision of Boole and his classical AI descendents. For example, although nearly all adults that we've tested can correctly define what makes a number even and claim that all three-sided polygons are triangles, they nevertheless often mistake numbers like 798 to be odd \citep[enough to sway a local election][]{hasen_wrong_2012} and systematically omit pictures of scalene triangles when choosing triangles. When asked why, some insisted that those were not \textit{really} triangles \citep{lupyan_difficulties_2013}. This graded pattern-matching nature of human thought is the subject of hundreds of studies \citep[e.g.,][]{goodwinAcquisitionBooleanConcepts2013, wason1968reasoning, pollardContentContextEffects1987, wangContextEffectsProduced2014, oakhillBelievabilitySyllogisticReasoning1989} which collectively show that human reasoning is neither haphazard nor normatively logical, but is best understood as a form of pattern matching \citep{margolisPatternsThinkingCognition1990}.

The great irony is that while Boole thought he was discovering the laws of thought of human thinking, implementing these laws in machines (i.e., computers) was so useful precisely because these machines could do what we struggle with: compute logical relations independent of content (i.e., carry out the operation of A OR NOT B independently of the meaning of A and B). To put in more prosaic terms: While our arithmetic abilities are both slow and pattern-driven \citep{banerjeeChildrensArithmeticSkills2025, oshaughnessyDiverseMathematicalKnowledge2023}, a calculator that can add 3+5, can also add 5+3; its adding algorithm is blind to the values being added.

Although computers running programs that strictly separate content from process have transformed our world, the effort to implement domain-general artificial intelligence through such programs has not succeeded. How ironic then that the breakthroughs that led to modern LLMs came from eschewing principles originally formulated to describe human thought, in favor of systems that come to perform the kind of messy probabilistic pattern matching that seems to underlie actual human thought! It is only by thoroughly collapsing content and process that it becomes possible to make sense of \textit{He dwushed a ghanc zawk}.

Of course we should not ignore that our species has discovered ways of overcoming some of the limitations of a pattern-matching brain. How do we explain our ability to do math and science if thinking is just messy pattern-matching? Some continue to insist that despite the evidence to the contrary, \textit{real} thinking operates on Boolean or other algebraic principles \citep{marcusHowO3Grok2025}. We think a more promising approach lies in studying how our pattern-matching minds are extended by cognitive prostheses which allow us to formulate and manipulate progressively more abstract and larger patterns. These include numeral systems and mathematical notations, maps, diagrams, clocks \citep{heyesCognitiveGadgetsCultural2018, hutchins2000distributed, tverskyCognitiveDesignTools2015, chrisomalisReckoningsNumeralsCognition2020}, and importantly, language itself \citep{clarkMaterialSymbols2006, lupyan_centrality_2016, lupyan_how_2016}.

In the paper that inspired the title of this article, the physicist Eugene Wigner \citep{zotero-item-20956} remarked on the miracle that allows us to derive physical laws from mathematical theory. Wigner compared this “unreasonable effectiveness of mathematics” to the “miracle that the human mind can string a thousand arguments together without getting itself into contradictions.” Putting aside the quibble that our minds can and often do get tangled in contradictions, the real miracle may be how far pattern matching can take us.

\section{Acknowledgments}
This work was partially funded by NSF-PAC 2020969 to the first author. We'd like to thank Senyi Yang for developing code to test Jabberwockified translations at scale.

\clearpage
\appendix

\section{Appendix}
\subsection{Example of an LLM playing \textit{The Gostak}.}
\label{subsec:gostak}

The system prompt instructs the model to reflect on the situation, articulate its thinking, and issue a command which is fed to the game engine. The game’s output is sent back to the model.

\begin{quote}
>SITUATION: I am in the delcot of tondam, a place where doshes deave. Five glauds are crenned in a loff lutt. A gomway deaves to kiloff and kirf, gombing a samilen to its hoff.\\
THOUGHT: "Oh, glauds in the loff lutt. I wish I could pell back to the bewl and distunk them. But as a gostak, only I distim the doshes."\\
COMMAND: Inspect glauds.\\
GAME OUTPUT: That's not a dape I recognise.

>SITUATION: The game does not recognize the command I gave. I'm still at the delcot of tondam where five glauds are crenned in the loff lutt.\\
THOUGHT: "Hmm, seems like 'inspect' doesn't work. Maybe a more general command like 'look' will yield more information."\\
COMMAND: Look glauds.\\
GAME OUTPUT: That's not a dape I recognise.

>SITUATION: Still in the delcot of tondam, the game doesn't recognize my commands to interact with the glauds.\\
THOUGHT: "Perhaps I need to use the game-specific words. Could 'vorl' be the word to interact with the glauds?"\\
COMMAND: Vorl glauds.\\
GAME OUTPUT: Which do you mean, the raskable glaud, the poltive glaud, the glaud-with-roggler, the glaud of jenth or the Cobbic glaud?

>SITUATION: I used 'vorl' and the game asks me to specify which glaud to interact with: the raskable glaud, the poltive glaud, the glaud-with-roggler, the glaud of jenth, or the Cobbic glaud.\\
THOUGHT: "Interesting, each glaud seems unique. The 'gluad-with-roggler' sounds intriguing, I should interact with it."\\
COMMAND: Vorl glaud-with-roggler.\\
GAME OUTPUT: Vorling is the fesh of ghelipers.\\
A gitch tunks you from the hoggam.

>SITUATION: After I tried to 'vorl' the glaud-with-roggler, I'm told that vorling is the fesh of ghelipers. Then, a gitch tunks me from the hoggam.\\
THOUGHT: "Vorling seems to be an irrelevant action here. I got tuniked by a gitch, maybe I could try to tunik it back?"\\
COMMAND: Tunik gitch.\\
GAME OUTPUT: That's not a dape I recognise.

>SITUATION: The game does not recognize my command to 'tunik' the gitch. I am still in the delcot of tondam, with the glauds in the loff lutt.\\
THOUGHT: "Hmm, tuniking didn't work. Maybe I should try to distim the gitch, as a gostak, and see what occurs."\\
COMMAND: Distim gitch.\\
GAME OUTPUT: You distim one, but it frikes fosken.
\end{quote}

\newpage
\textbf{Elicited definitions of some Gostakian words after ~10 minutes of LLM gameplay}

\begin{table}[H]
\centering
\begin{tabular}{p{0.22\textwidth} p{0.68\textwidth}}
\toprule
\textbf{Word} & \textbf{Definition} \\
\midrule
\textbf{Dape} & Refers to a recognized or valid command. \\
\textbf{Reb} & Appears to mean `do` or `perform`. \\
\textbf{Tophthed} & Likely refers to a state of being trapped or incapacitated. \\
\textbf{Distim} & Refers to activating or interacting with objects (e.g., glauds). \\
\textbf{Distunk} & A variant of distim, possibly a more specific or focused interaction. \\
\textbf{Zank} & An action directed toward neutralizing or affecting entities (e.g., warbs or gitches). \\
\textbf{Frike} & Appears to be an action related to interaction with gitches. \\
\textbf{Kirf} & An action likely related to movement or navigating the environment. \\
\textbf{Gomb} & Refers to performing an action on an object or entity, possibly `engage` or `attempt`. \\
\textbf{Leil} & To interact with or trigger something, likely an object like a gropple. \\
\textbf{Shoracle} & A state or condition of gropples when interacted with, indicating activation or alteration. \\
\textbf{Jirf} & A key action or movement used to progress or navigate, possibly meaning `move forward` or `navigate`. \\
\textbf{Tunk} & An action associated with interaction, likely meaning `strike` or `engage`. \\
\textbf{Gostak} & Represents the player-controlled entity, potentially the protagonist or agent. \\
\bottomrule
\end{tabular}
\end{table}

\newpage
\subsection{An example of an LLM recovering text that has had all content words completely masked}

\label{subsec:totally_masked}
\vspace{2\baselineskip}
\noindent{\textbf{Original Text:}}

\begin{center}
  \begin{minipage}{0.7\textwidth}
    \ttfamily
In the United States, federal law has priority over any particular state’s law. If a state law conflicts with federal law, the state law is invalidated and the federal law is followed. For example, one federal law specifies what information should be included on the labels of medical devices. When Massachusetts passed a law that required additional information on hearing aid labels, federal courts barred Massachusetts from enforcing this law.\\
  \end{minipage}
\end{center}
\normalfont

\noindent\textbf{Prompt to Gemini 3 Pro:}\\

\begin{center}
  \begin{minipage}{0.7\textwidth}
    \ttfamily
    In this passage, content words have been replaced with the word BLANK. Translate the passage as best as you can into standard English. The final response should have all English words and retain consistency with the original text.\\

In the BLANK BLANK, BLANK BLANK has BLANK over any BLANK BLANK’s BLANK . If a BLANK BLANK BLANKs with BLANK BLANK , the BLANK BLANK is BLANKed and the BLANK BLANK is BLANKed. For BLANK , BLANK BLANK BLANK BLANKs what BLANK should be BLANKed on the BLANKs of BLANK BLANKs . When BLANK BLANKed a BLANK that BLANKed BLANK BLANK on BLANK BLANK BLANKs , BLANK BLANKs BLANKed BLANK from BLANKing this BLANK.
  \end{minipage}
\end{center}
\normalfont

\noindent\textbf{\href{https://gemini.google.com/share/583a50296c3f}{Output}}\\

\begin{center}
  \begin{minipage}{0.7\textwidth}
    \ttfamily

In the \textbf{United States}, \textbf{federal law} has \textbf{precedence} over any \textbf{state government}’s \textbf{authority}. If a \textbf{state law conflicts} with \textbf{federal law}, the \textbf{federal law} is \textbf{upheld} and the \textbf{state law} is \textbf{overturned}. For \textbf{example}, \textbf{federal law strictly regulates} what \textbf{warnings} should be \textbf{placed} on the \textbf{labels} of \textbf{cigarette packages}. When \textbf{Massachusetts passed} a \textbf{law} that \textbf{required graphic warnings} on \textbf{cigarette package labels}, \textbf{federal courts blocked Massachusetts} from \textbf{enforcing} this \textbf{legislation}.
  \end{minipage}
\end{center}

\normalfont

\clearpage
\printbibliography
\clearpage

\end{document}